\documentclass{article}
\usepackage{amsmath,graphicx}
\usepackage[square, comma, sort&compress, numbers]{natbib}

\title{Deep Steganalysis:  End-to-End Learning with Supervisory Information beyond Class Labels}
\date{}
\author{Wei~Wang,      Jing~Dong,      Yinlong~Qian,    and~Tieniu~Tan}

\begin{document}
\maketitle
\begin{abstract}
Recently, deep learning has shown its power in steganalysis. However, the proposed deep models have been often learned from pre-calculated noise residuals with fixed high-pass filters rather than from raw images. In this paper, we propose a new end-to-end learning framework that can learn steganalytic features directly from pixels. In the meantime, the high-pass filters are also automatically learned. Besides class labels, we make use of additional pixel level supervision of cover-stego image pair to jointly and iteratively train the proposed network which consists of a residual calculation network and a steganalysis network. The experimental results prove the effectiveness of the proposed architecture.
\end{abstract}
\section{Introduction}
Steganography is the technique of concealing communication by means of cover medium transmission. It is a hot topic of information security and has drawn more and more attention in recent years. On the contrary, steganalysis aims at detecting the very existence of secret message in covers such as digital images. It is a very challenging task because the stego signal is usually very weak while greatly impacted by the variations of cover contents. Generally, it can be seen as a binary classification problem that distinguishes stegos from covers. For a long time, detection is conducted with two separate steps: feature extraction and classification. It would be worth to note that, though classifier is automatically optimized, the crucial feature extraction is based entirely on heuristics. To detect the advanced steganographic schemes \cite{pevny2010using,holub2012designing,holub2013digital}, various handcrafted high-dimensional features such as rich feature representations \cite{gul2011new,fridrich2011steganalysis,fridrich2012rich,shi2013textural,holub2013randomJ,tang2014adaptive,denemark2015selection, ensemble_classifier} have been proposed, even though the image is hard to modeled.

In recent years, based on the concept of learning features automatically, deep learning has made significant achievements in various areas like computer vision. The use of deep learning challenges traditional handcrafted feature based approaches. This has aroused the interest of researchers in the steganalysis field in seeking the way to apply deep learning for steganalysis. In our very first paper \cite{qian2015deep}, using deep convolutional neural networks (CNN) with high-pass filtering layer, we achieve comparable performance with that of the SRM\cite{fridrich2012rich} with ensemble classifiers \cite{ensemble_classifier}. Inspired by this paper, different architectures are proposed. In \cite{pibre2016deep}, Pibre et al. present a CNN architecture for steganalysis with fewer convolutional layers, and without the pooling operation. But such architecture is designed specially for steganalysis in the scenario where the embedding happened roughly in the same locations over images caused by a fixed embedding key. This is a similar situation for the architecture proposed in \cite{Couchotsteg16}. More recently, Xu et al. also propose a deep learning approach for steganalysis with a more compact CNN model \cite{xu2016structural} and some ensemble strategies \cite{xu2016ensemble} that have led to performance improvement. Furthermore, we pre-train a CNN model with high payload stego images and fine-tune it with lower payload ones \cite{Qian2016transfer}. The effectiveness of this transfer learning strategy has been proven. Another effective transfer learning strategy incorporating the prior knowledge from traditional steganalysis tasks to regularize a CNN model has been proposed in \cite{Qian2016reg}.

So far, deep learning shows its power in steganalysis. However, most of the proposed models are learned from pre-calculated noise residuals rather than raw images. In fact, CNNs are originally designed to learn patterns for understanding visual content of the image in computer vision problems, while in steganalysis, the signal of interest is hidden within the noise component. In order to deal with this problem, recent methods firstly filter the image with a handcrafted high-pass kernel to enhance the stego signal, and then use CNN to learn features from computed residuals. The limitation is that handcrafted kernels could be suboptimal when facing sophisticated data embedding methods and image sources. In this paper, we propose a new CNN architecture that can learn the residuals and the steganalytic features from images simultaneously. Actually, it is hard to learn such residuals with only label information. We make use of cover-stego image pairs as additional supervisory information to train the proposed network jointly and iteratively. The experimental results prove the effectiveness of the proposed method. 

\section{The Proposed Method}
The core idea behind deep learning is that comprehensive feature representations can be efficiently learned with the deep architectures which are composed of stacked layers of trainable non-linear operations. However, because of the diversity of image content, it is hard to learn effective feature representations directly from images for steganalysis. Recent proposed approaches have to fix the kernel of first layer as the HPF (high-pass filter) like the kernel in \cite{fridrich2012rich}. It is so-called pre-processing layer. In other words, only label information is not enough to learn powerful feature representations for steganalysis. In this section, we present a novel framework based on deep learning with more supervisory information beyond class labels. 

\subsection{Overall Architecture}

As we said above, because of the power of deep networks and large-scale training data, deep learning always gets much better performance than traditional learning methods. However, for steganalysis, it is hard to say that since image content instead of stego signals dominates the optimization process. To tackle this problem, In \cite{qian2015deep}, we propose a feasible approach by employing the HPF kernel as defined in Eq. (\ref{kv}) at the first layer of the proposed network. We fix the parameters without learning at this layer.
\begin{equation}\label{kv}
\textbf{K}_{5\times5}=\frac{1}{12}\left(
  \begin{array}{rrrrr}
    -1 &  2 & -2 &  2 & -1\\
     2 & -6 &  8 & -6 &  2\\
    -2 &  8 &  -12 &  8 & -2\\
     2 & -6 &  8 & -6 &  2\\
    -1 &  2 & -2 &  2 & -1 
  \end{array}
\right)
\end{equation}
The role of the first layer of fixed HPF is to suppress the image content while amplifying the stego signals, which can make the deep networks learn powerful feature representations for steganalysis. Thereafter, different networks have been proposed \cite{Qian2016transfer, Qian2016reg, xu2016structural, xu2016ensemble} but with the same fixed HPF layer.

Can we learn this HPF automatically or, in other words, can we learn steganalytic features directly from an image itself? Since stego signal is quite weak and its existence is easy to be concealed by the variety of image contents, we found that it is hard to directly learn HPF with only label supervision (cover or stego). But we have more than that. Cover and stego image pair as the supervisory information at the pixel level is also used. Firstly, we reconstruct the cover images using this pixel level supervisory information. The idea of image reconstruction is similar to cover estimation in traditional steganalysis. But differently, we make this step trainable. Secondly, by subtracting reconstructed image contents from input images, noise residuals are obtained.  It is worth noting that the learned filters in the image reconstruction step are low-pass, but the whole two-step module can be recognized as high pass filtering.  

Generally, steganalysis is the binary classification problem that optimizes over a loss function. In our proposed approach, we formulate the loss function  as a combination of two terms: the classification term and the image reconstruction term. The classification term is a global constraint that makes the prediction accurate, and the reconstruction term reconstructs the input image at the pixel level, as seen in Eq. (\ref{loss}). 
\begin{equation}\label{loss}
    J(\theta_c,\theta_r)=( {1 - \lambda })J_c(\theta_c) + \lambda J_r(\theta_r),
\end{equation}
where $\lambda \in [0,1]$ specifies a relative importance of the reconstruction term $J_r$ against the classification term $J_c$. The goal of the reconstruction term is to ensure that the noise residuals can be automatically learned from the input image.

\begin{figure*}
\centering
\includegraphics[width=0.8\textwidth]{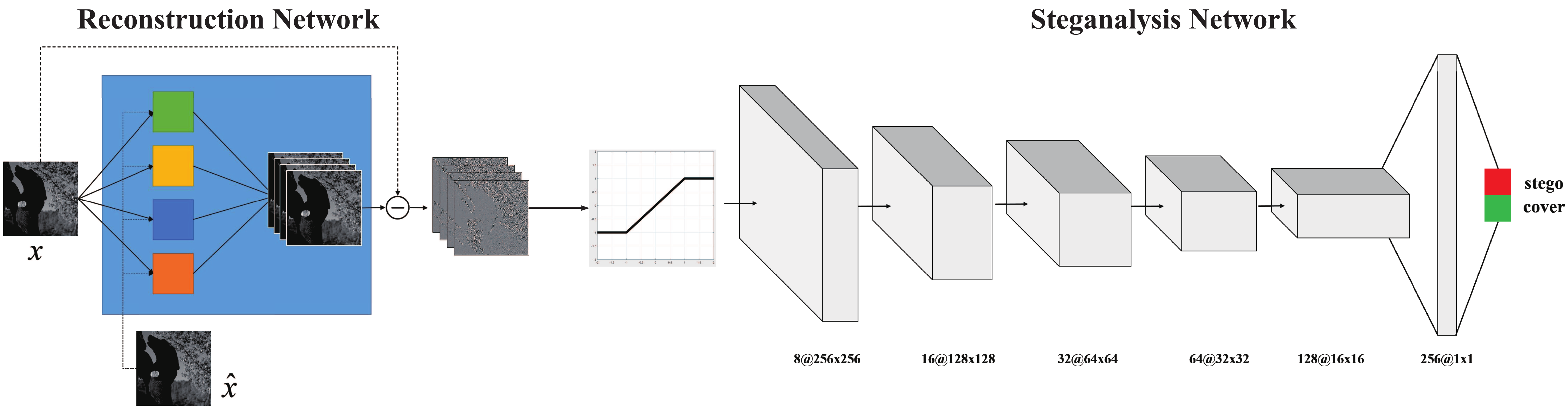}
\caption{The overview of our proposed end-to-end architecture which has two sub-networks: the residual network and the steganalysis network. The residual network optimally reconstructs the input image $x$ at the pixel level and outputs residuals as the input of the steganalysis network. $\hat{x}$ is the corresponding cover image of $x$.}
\label{fig:overall}
\end{figure*}

The overall architecture is shown in Fig. \ref{fig:overall}. It has two sub-networks: the residual network and the steganalysis network. The residual network is placed at the very beginning to transform original images to its reconstructions, then to get noise residuals. The residuals are calculated as the differences between the input image and its reconstructions straightforwardly. Then, we feed the residuals into the steganalysis network. In the following, we present the details of the two sub-networks and their corresponding loss functions. 

\subsection{The Residual Network}
Image reconstruction is the key part of the residual network. A straightforward way to learn the reconstruction model is to train a regression network $r$ to make the response $r(x)$ of an input image $x$ approximate the groundtruth which should be the corresponding cover image of $x$. As we all know, the loss criterion plays an important role in learning process. For the residual network, we define its reconstruction loss as 
\begin{equation}\label{loss2}
    J_r(\theta)=\frac{{\rm{1}}}{n}\sum\limits_{i = 1}^n {\Omega(r({x^i};\theta_r),\hat{x}^i)}
\end{equation}
in which, supposing $x^i$ has a total of $m$ elements, 
\renewcommand{\arraystretch}{1.5}    
\begin{equation}\label{reconloss}
  \Omega (x,{\hat x}) = \frac{1}{m} \sum \left\{ 
    \begin{array}{{ll}}
        {\frac{1}{2}({x_j} - {\hat x}_j)}^2 & {if\text{   $|{{x}_j - {{\hat x}}_j}| < 1$}}\\
        {|{{x}_j - {{\hat x}}_j}| - \frac{1}{2}} & {otherwise}
    \end{array}\right.
\end{equation}
is a robust $L1$ loss that is less sensitive to outliers than the $L2$ loss, when $|{{x}_j - {{\hat x}}_j}| \ge 1$. Meanwhile, when $|{{x}_j - {{\hat x}}_j}|$ is quite close to zero, $L2$ loss is used to dampen oscillation. It should be noted that the input images are normalized. $\hat{x}^i$ is the corresponding cover image of the $i$th input image $x^i$, and $r({x^i};\theta)$ is the reconstructed image. $n$ is the number of training samples.

\begin{figure}[!htbp]
\centering
\includegraphics[width=0.4\textwidth]{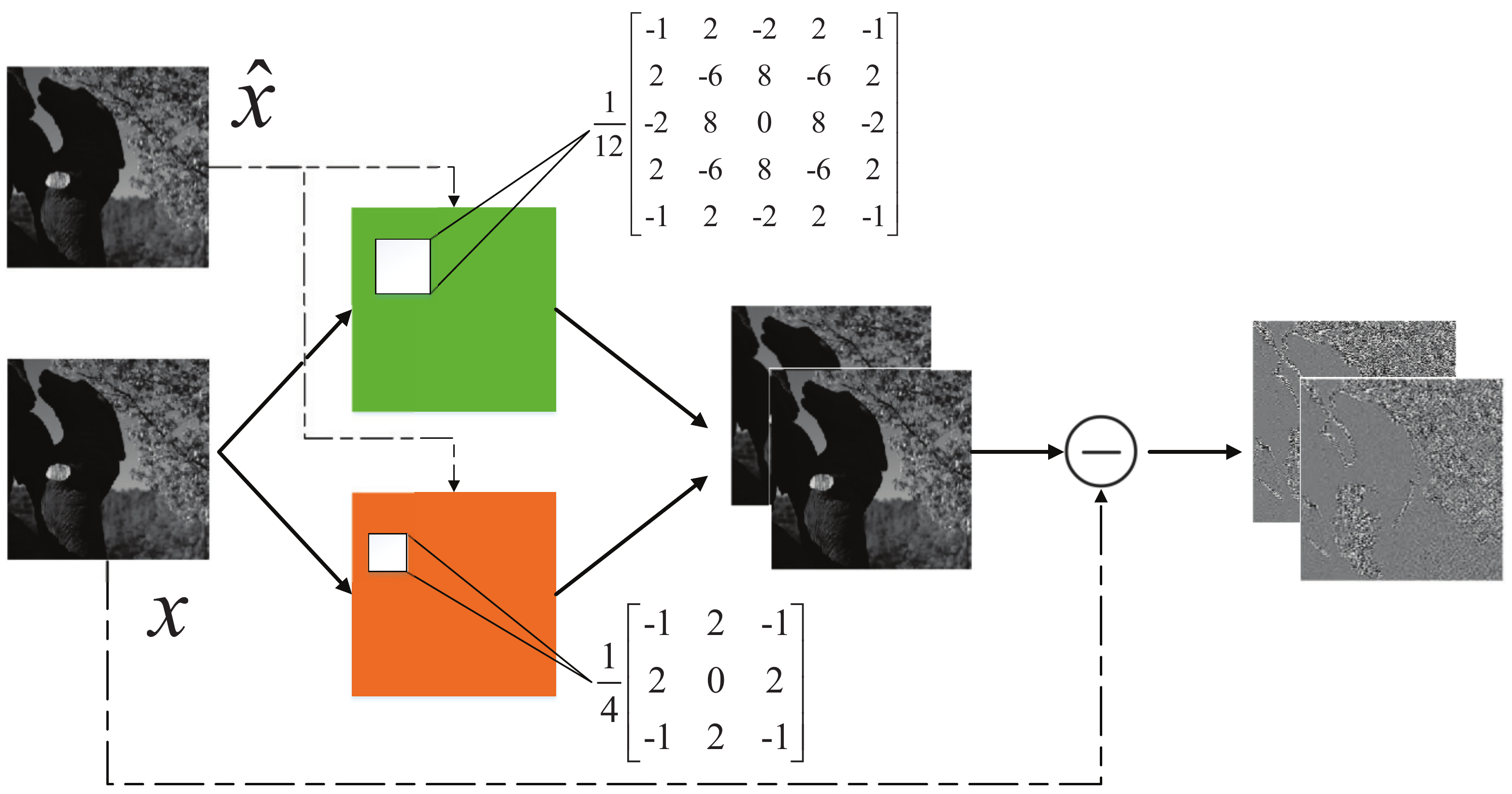}
\caption{The details of the residual network.}
\label{fig:recon}
\end{figure}

We design a one convolutional layer network to approximate $r$ because more layers means more information loss, which is fatal for stego signal that we try best to preserve. The residual noise is calculated as the difference between the input image $x^i$ and its reconstruction $r(x^i)$ straightforwardly. Fig. \ref{fig:recon} shows the architecture of residual network. It has two learnable kernels with sizes of $5\times5$ and  $3\times3$ respectively. As we know, parameter initialization is very important for training the networks. These two kernels are initialized with the weights as following. 
\begin{equation*}
\small
\frac{1}{12}\left(
  \begin{array}{ccccc}
    -1 &  2 & -2 &  2 & -1\\
     2 & -6 &  8 & -6 &  2\\
    -2 &  8 &  0 &  8 & -2\\
     2 & -6 &  8 & -6 &  2\\
    -1 &  2 & -2 &  2 & -1
  \end{array}
\right),
\text{ and }
\frac{1}{4}\left(
  \begin{array}{ccc}
    -1 &  2 & -1 \\
     2 & 0 &  2 \\
    -1 &  2 & -1
  \end{array}
\right)
\end{equation*}
We train the network in a supervised way with the corresponding covers guiding the training process. It should be noted that the reconstruction loss is the average of losses corresponding to the two outputs. The outputs are the reconstructed images, respectively

Then by taking the differences between the input and the outputs, we get two residuals which are finally fed into the steganalysis network that will be introduced in Section \ref{sec:stegnet}

\subsection{The Steganalysis Network}
\label{sec:stegnet}
For the stegnalsysis network, it learns feature representation automatically and optimally. The objective is to minimize the classification loss function defined in Eq. (\ref{loss1}). 
\begin{equation}\label{loss1}
    J_c(\theta_c)=\frac{{\rm{1}}}{n}\sum\limits_{i = 1}^n {L(f({x^i};\theta_c ),{y^i})},
\end{equation}
where $f$ is the predicted probability distribution over different possible outcomes (cover or stego), and $L$, similar to the loss function used in logistic regression, is the negative log-likelihood of the correct class,
\begin{equation}\label{ClassNLLCriterion}
L(x,y;\theta_c) = -y \log f(x;\theta_c ) - (1-y) \log (1-f(x;\theta_c )),
\end{equation}
where $x$ is the input residuals and $y$ is the corresponding label.
In the proposed architecture, we use the Xu's\cite{xu2016structural} network as the steganlysis network to approximate $f$. The models proposed by us in \cite{qian2015deep} and by Xu et al. in \cite{xu2016structural} are basically the same except the additional abs layer and batch normalization layers which make Xu's Net have less parameters and consequently less probability of over-fitting.

\subsection{End-to-End Learning}
Since the reconstructed image can be learnt optimally, assembled with the steganalysis network, the proposed steganalytic CNN model can directly learn from raw images.

We iteratively update the parameters of the proposed architecture by minimizing the joint loss (\ref{miniloss}) through stochastic gradient descent (SGD). $\lambda$ is set to $0.999$ in our implementation. During the optimization, the parameters of the residual network are updated using the gradients both from the residual network itself and the steganalysis network. 
\begin{equation}\label{miniloss}
    \mathop {\arg \min }\limits_{{\theta _c},{\theta _r}}J(\theta_c,\theta_r) 
\end{equation}
Both the corresponding cover image and class label are used as the supervisory information to guide the learning process.


\begin{figure}
\centering
\includegraphics[width=0.4\textwidth]{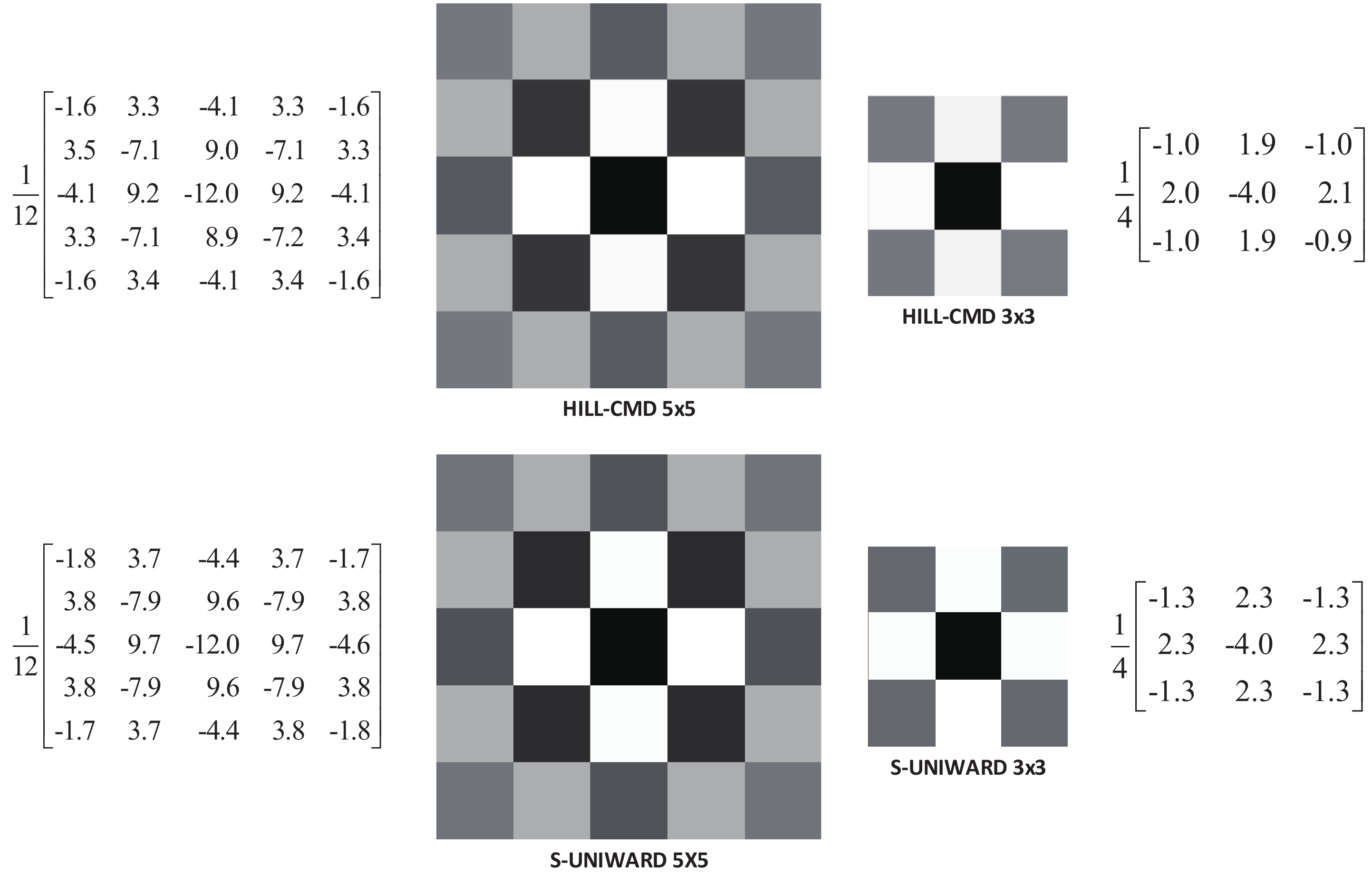}
\caption{The learned high-pass filters of our proposed architecture \emph{Recon-Steg-Net} on detecting S-UNIWARD and HILL-CMD with embedding rate 0.4bpp.}
\label{fig:hpf}
\end{figure}

\begin{figure*}
\centering
\includegraphics[width=0.9\textwidth]{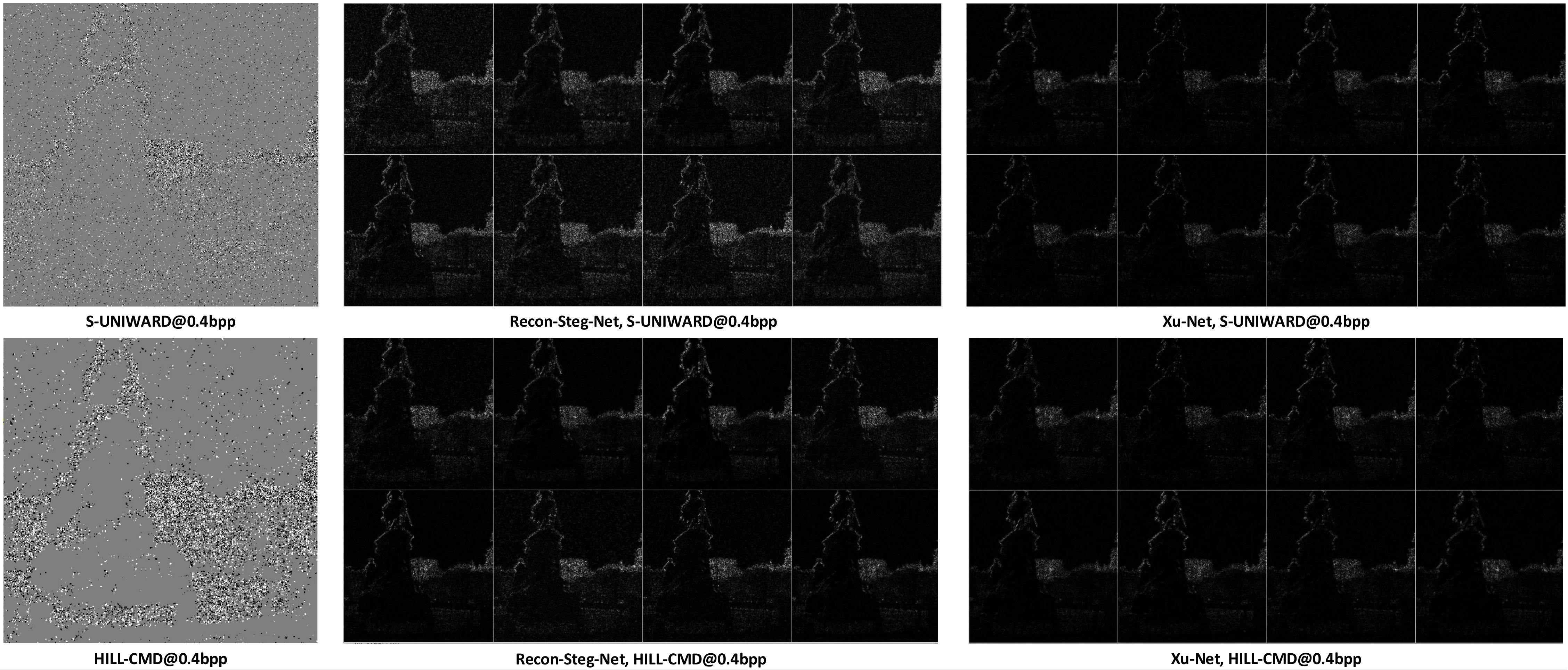}
\caption{The feature maps of `abs layer' in our proposed (middle) and Xu's (right) models on detecting S-UNIWARD and HILL-CMD with embedding rate 0.4bpp. The left colum shows the embedding changes.}
\label{fig:abslayer}
\end{figure*}

\section{Experiments}
\subsection{Dataset}
The dataset we used is BOSSbase v1.01 \cite{bas2011break} which contains 10000 images acquired by seven digital cameras in raw format and subsequently processed to the size of $512\times512$. To evaluate the performance of our proposed CNN models, four state-of-the-art spatial domain content-adaptive steganographic algorithms,  S-UNIWARD \cite{holub2013digital}, HILL-CMD \cite{li2015strategy}, WOW\cite{holub2012designing}, and MiPOD\cite{sedighi2016content} with embedding rate of 0.4 bpp are considered. They were implemented with unfixed random embedding keys. 
\subsection{Parameter Configuration}
 The stochastic gradient descent (SGD) algorithm was used to train the proposed CNN in experiments. The learning rate was initialized to 0.001 and scheduled to lower it by a multiplier of 0.3 for five times when the error plateaus lasts for some amount of time (50 epochs). The momentum was set to 0.9.  The minibatch size was 32 (16 cover/stego pairs) because of the limitation of GPU memory. All the weights in convolutional layers are initialized from a zero-mean Gaussian distribution with standard deviation of 0.01. Bias learning was disabled in the convolutional layers. The weights in the last fully connected (FC) layer were initialized using ``Xavie" initialization and the bias were initialized with $0$. The weight decay was not enabled except that for the FC layer it was fixed to 0.0005.

\subsection{Results}
We compared the performance of proposed CNN model with the state-of-the-art CNN model proposed by Xu et al. \cite{xu2016structural}. They are both implemented with Torch under the same parameter configuration. Since we mainly focus on the possibility of learning deep steganalytic model from a raw image, no ensemble strategy was adopt for both models. In the experiments, 8000 training pairs (cover/stego) were randomly selected from the dataset. The remains were used as testing pairs which were never touched in the whole training phase. In the training phase, we randomly generated a split from the 8000 training pairs by evenly breaking the 8000 training pairs into five non-overlapping folds, and used the first four folds for training and the rest one fold for validation. The models were trained on 6400 pairs and validated on 1600 pairs. The validation pairs were used to decide when to lower the learning rate or stop the training process. To evaluate the performance, in the testing stage, the 2000 testing pairs went through the well-trained models. Table \ref{tab:err} shows the detection error rates. 
We can find that our proposed model has better performance on detecting MiPOD, WOW and HILL-CMD, and also is comparable to Xu's model when detecting S-UNIWARD. This proves the effectiveness of our proposed model. That is to say we can directly learn effective steganalytic features from images themselves without using the fixed HPF layer.

\begin{table}
\renewcommand{\arraystretch}{1.2}
\caption{Detection Error rates on the state-of-the-art steganography methods}
\vspace{1em}
\label{tab:err}
\centering
\resizebox{0.9\linewidth}{!}{
\begin{tabular}{c||c|c|c|c} 
 \hline
  & S-UNIWARD & HILL-CMD & WOW & MiPOD\\
 \hline
 Xu's & 20.68\% & 30.83\% & 19.53\% & 24.78\% \\
 \hline
 Proposed & 20.78\% & \textbf{29.35\%} & \textbf{17.78\%} & \textbf{22.43\%} \\
 \hline
\end{tabular}
}
\end{table}

\subsection{Visualization}
Fig. \ref{fig:hpf} shows the normalized learned high-pass filters and their visualization for HILL-CMD and S-UNIWARD algorithms.  We can find that the learned HPF filters have similar shapes to the traditionally used $5\times5$ and $3\times3$ filters, but have more  weights on surrounding pixels which makes the stego-noise more exposed to the corresponding residuals. This can be verified in Fig. \ref{fig:abslayer} especially in the so-called hard-to-predicate texture regions. Fig. \ref{fig:abslayer} shows the feature maps after 'abs layer' in steganalysis network \cite{xu2016structural}. We can find that our proposed residual network makes the steganalysis network capture more details on texture region, which is very important for detecting adaptive steganography algorithms. 

\section{Conclusion}
In this paper, we proposed to learn effective steganalytic feature representations directly from images themselves. We made use of both label information and cover-stego pair information to train the proposed network by minimizing the joint loss (\ref{miniloss}). Results proved that the high-pass filters can be optimally learned. The proposed model captured more details on texture region, which boosts the detection performance especially for MiPOD, WOW and HILL-CMD. In our future work, we will put more effort on learning more types of high-pass filters to improve the performance. How to incorporate the knowledge of the selection-channel into deep models is also an interesting topic.

\bibliographystyle{unsrt}

\end{document}